\pdfoutput=1
\documentclass[conference]{IEEEtran}
\IEEEoverridecommandlockouts 

\usepackage{microtype}
\usepackage{amsmath,amssymb,amsfonts}
\usepackage{graphicx} \graphicspath{ {figures/} }
\usepackage{tabularx}
\usepackage{booktabs}
\usepackage{floatrow} 
\usepackage{balance}
\usepackage[colorlinks,pagebackref]{hyperref}

\DeclareMathOperator{\concat}{concat}
\DeclareMathOperator{\softmax}{softmax}

\begin{document}

\title{CFSum: A Transformer-Based Multi-Modal Video Summarization Framework With Coarse-Fine Fusion
}

\author{\IEEEauthorblockN{Yaowei Guo\IEEEauthorrefmark{1}}\thanks{* Corresponding author}
\IEEEauthorblockA{\textit{University of California, Los Angeles} \\
yaowei@cs.ucla.edu}
\and
\IEEEauthorblockN{Jiazheng Xing}
\IEEEauthorblockA{\textit{Zhejiang University} \\
jiazhengxing@zju.edu.cn}
\and
\IEEEauthorblockN{Xiaojun Hou}
\IEEEauthorblockA{\textit{Zhejiang University} \\
xiaojunhou@zju.edu.cn}
\and
\IEEEauthorblockN{Shuo Xin}
\IEEEauthorblockA{\textit{Zhejiang University} \\
22232036@zju.edu.cn}
\and
\IEEEauthorblockN{Juntao Jiang}
\IEEEauthorblockA{\textit{Zhejiang University} \\
juntaojiang@zju.edu.cn}
\and
\IEEEauthorblockN{Demetri Terzopoulos}
\IEEEauthorblockA{\textit{University of California, Los Angeles} \\
dt@cs.ucla.edu}
\and
\IEEEauthorblockN{Chenfanfu Jiang}
\IEEEauthorblockA{\textit{University of California, Los Angeles} \\
cffjiang@ucla.edu}
\and
\IEEEauthorblockN{Yong Liu}
\IEEEauthorblockA{\textit{Zhejiang University} \\
yongliu@iipc.zju.edu.cn}
}

\maketitle

\begin{abstract}
Video summarization, by selecting the most informative and/or user-relevant parts of original videos to create concise summary videos, has high research value and consumer demand in today's video proliferation era. Multi-modal video summarization that accomodates user input has become a research hotspot. However, current multi-modal video summarization methods suffer from two limitations. First, existing methods inadequately fuse information from different modalities and cannot effectively utilize modality-unique features. Second, most multi-modal methods focus on video and text modalities, neglecting the audio modality, despite the fact that audio information can be very useful in certain types of videos. In this paper we propose CFSum, a transformer-based multi-modal video summarization framework with coarse-fine fusion. CFSum exploits video, text, and audio modal features as input, and incorporates a two-stage transformer-based feature fusion framework to fully utilize modality-unique information. In the first stage, multi-modal features are fused simultaneously to perform initial coarse-grained feature fusion, then, in the second stage, video and audio features are explicitly attended with the text representation yielding more fine-grained information interaction. The CFSum architecture gives equal importance to each modality, ensuring that each modal feature interacts deeply with the other modalities. Our extensive comparative experiments against prior methods and ablation studies on various datasets confirm the effectiveness and superiority of CFSum.
\end{abstract}


\section{Introduction}

Hundreds of thousands of videos are uploaded to the various internet platforms every hour, and so it is becoming increasingly urgent to help users quickly identify videos they want to consume from the vast sea of content. Video summarization, the creation of concise summary videos that include the most informative parts of original videos, has emerged as a primary solution \cite{sen2019video, gygli2015video, ma2020similarity, zhang2016summary}. Video summarization also assists various industries in automating the management of videos.

\begin{figure}

\centering
\includegraphics[width=0.95\textwidth,height=0.5\textwidth]{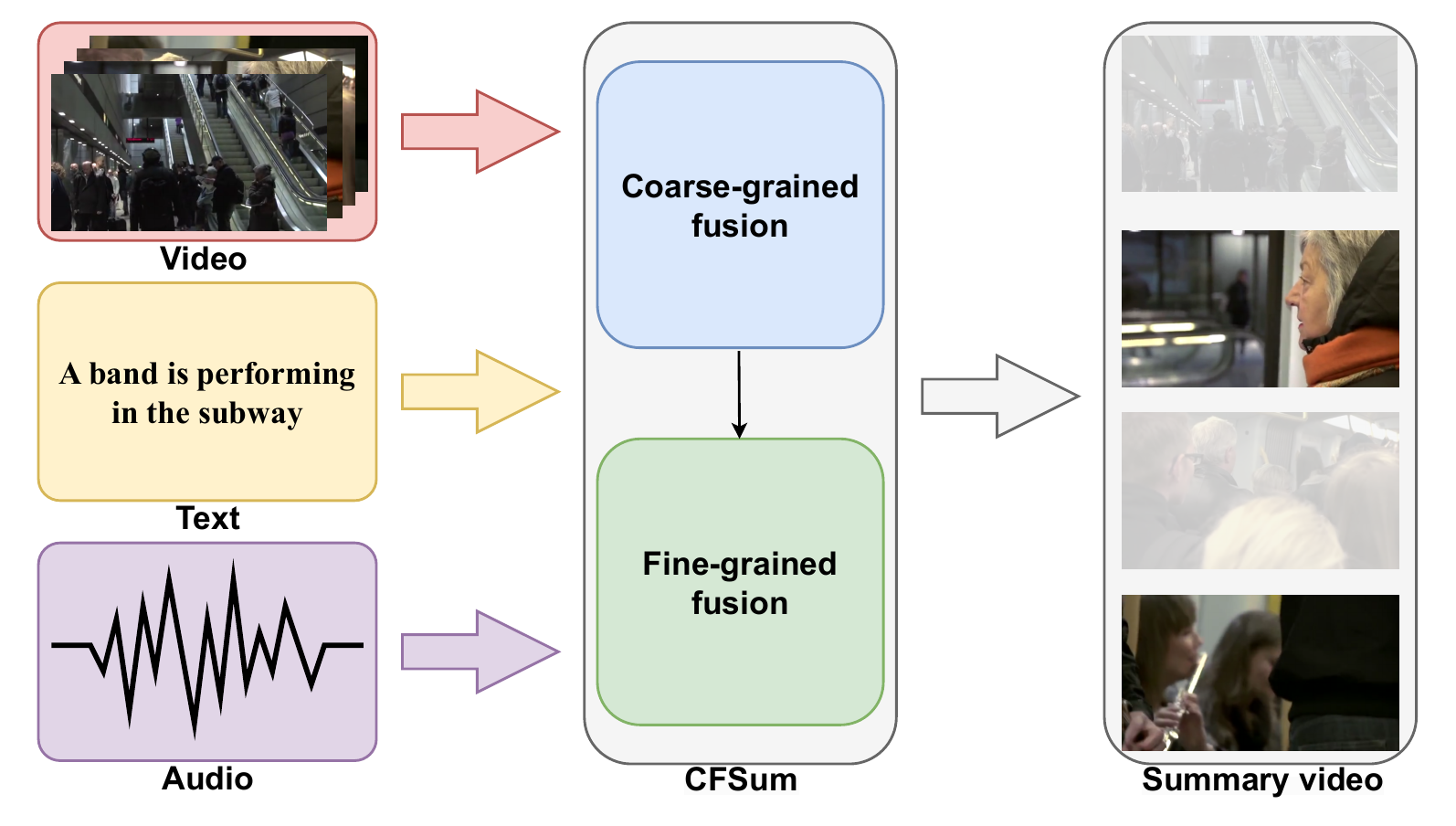}
\caption{Given video, audio, and text as inputs, CFSum leverages multi-modal coarse-fine fusion to generate a summary video by selecting from the original video the most important parts or those relevant to the user's input.}
\label{fig1}
\end{figure}

Multi-Modal Video Summarization (MMVS), which incorporates text input, has become a research focus in recent years \cite{liu2022umt, ma2023llavilo, moon2023query, narasimhan2021clip, narasimhan2022tl}. This approach allows users to input a segment of text, referred to as a \textit{query}, as a basis for generating the summary, which further enhances the capabilities and applicability of video summarization. However, existing MMVS methods have two major limitations: (1) These methods have insufficient fusion and interaction mechanisms for different modal features \cite{liu2022umt, moon2023query, park2022multimodal}; therefore, they fail to utilize effectively the unique information specific to each modality. Typically, they treat video input as \textit{key} and apply the text input as \textit{query} in transformer-based attention mechanism operations, and then incorporate video clips with high saliency scores into the summary video. (2) Most MMVS models consider only video and text as inputs \cite{ma2023llavilo, xiao2024bridging, narasimhan2021clip}; therefore, they neglect the unique features that the audio modality can provide. In certain types of videos, such as instructional videos, sports videos, etc., audio features contain richer and more prominent information compared to other modalities. 

\begin{figure*}[ht]
\centering
\includegraphics[width=0.8\textwidth,height=0.24\textwidth]{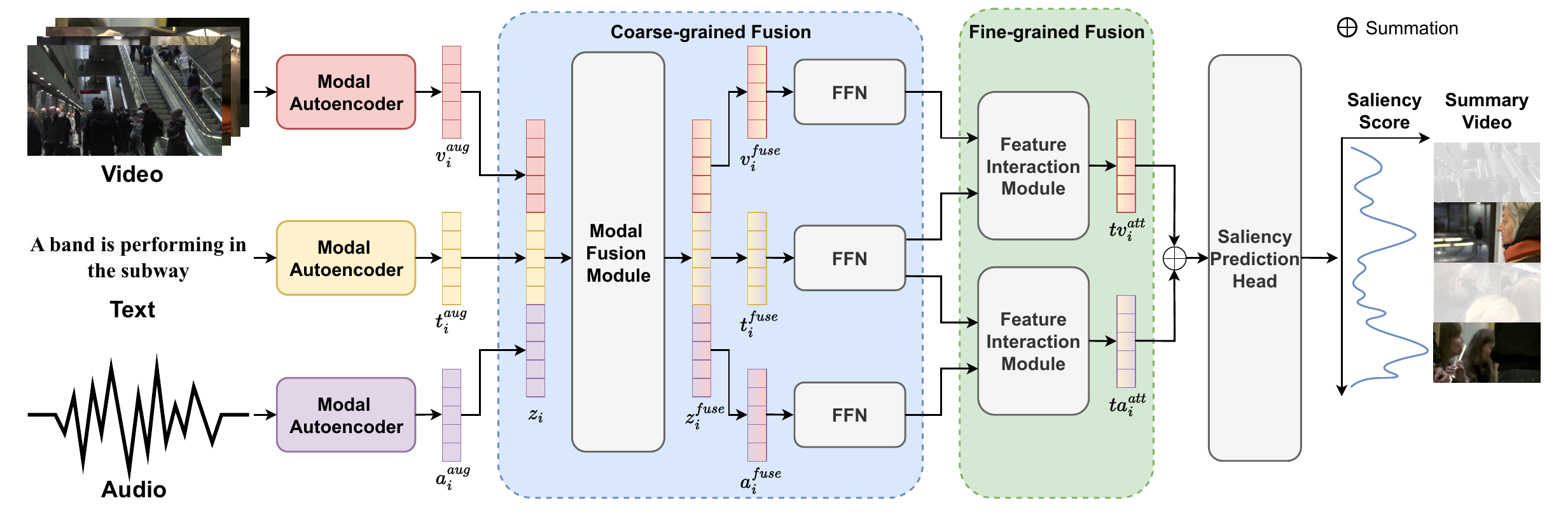}
\caption{CFSum model architecture. Upon receiving multi-modal inputs, CFSum first leverages modal autoencoders to enhance features by intra-modal learning. The results are then input to a modal fusion module that performs coarse-grained fusion. This is followed by feature interaction modules that perform fine-grained fusion. Finally, a saliency prediction head predicts the saliency scores. }
\label{fig2}
\end{figure*}

Motivated by the above considerations, we propose CFSum, a novel transformer-based MMVS framework with a coarse-fine fusion architecture, as illustrated in \autoref{fig1}. CFSum employs a two-stage modal fusion structure to achieve comprehensive feature fusion. In the first stage, to ensure that the feature representation of each modality incorporates information from other modalities, a modal fusion module uses self-attention to perform unified fusion of multi-modal features. In the second stage, to further facilitate feature interaction, feature interaction modules employ cross-attention operations between modalities to fully utilize modality-specific information and mitigate the impact of background noise, thereby achieving comprehensive modality interaction and fusion. By virtue of this two-stage fusion structure, CFSum achieves deep fusion both within and between modalities in order to generate a fully-fused video-audio-text representation. Finally, a saliency head predicts saliency scores to produce the summary video.

Hence, this paper makes the following major contributions:
\begin{itemize}
\item We introduce CFSum, a novel multi-modal video summarization framework featuring coarse-fine fusion that, unlike previous methods, achieves a comprehensive fusion of multi-modal features. 
\item In contrast to previous methods that ignore the audio modality, in CFSum we value the information contained in audio, elevating its importance to equal that of video and text.
\item We perform extensive experimentation on TVSum, Youtube Highlights and QVHighlights datasets to affirm the superiority of CFSum relative to multiple previous MMVS methods, and our detailed ablation studies examine the contributions of each module and modality across different video categories.
\end{itemize}

\section{Related Work}

Most MMVS methods focus on feature fusion between
modalities. Early ones focused on summarizing specific types of videos; e.g., Sanabria \emph{et al.} \cite{sanabria2019deep} used commentary to summarize sports videos, and Li \emph{et al.} \cite{li2017extracting} employed other sensor modalities to summarize first-person videos.

In recent years,
MMVS models based on visual object detection transformers have emerged \cite{liu2022umt, ma2023llavilo, moon2023query, narasimhan2021clip, narasimhan2022tl, xiao2024bridging, moon2023correlation, park2022multimodal, he2023align}. Among them, Narasimhan \emph{et al.} \cite{narasimhan2021clip} used a query-guided transformer that predicts frame saliency scores based on the significance between video frames and their relevance to input query statements; Liu \emph{et al.} \cite{liu2022umt} introduced the audio modality and designed a visual-text-audio multi-modal transformer-based framework featuring a query generator and query decoder to address the keyframe detection problem; and Moon \emph{et al.} \cite{moon2023query} enhanced model's ability to utilize textual query information by injecting into it negative sample video-text pairs.

A few MMVS models consider audio inputs (e.g., \cite{liu2022umt, moon2023query,  park2022multimodal}), but they either employ complex fusion methods that are computationally expensive, or they give audio insignificant importance, treating it merely as auxiliary information for text. Our CFSum model not only features a clear and concise multi-modal fusion method, but also effectively exploits the information characteristic to each modality.

\section{Method}


Video summarization aims to select the most informative or user-relevant parts of an original video to form a summary video. Formally, given a video $V$ comprising $n_c$ clips and a text query $T$ provided with $n_t$ tokens, the goal is to predict the saliency score of each clip $s_{i \in \{1,...,n_c\}}$ based on its importance or relevance to the text query.

\autoref{fig2} shows the overall architecture of our CFSum model. It consists of four parts: a modal autoencoder, a modal fusion module, a feature interaction module, and a saliency prediction head. Given the input video features $v_{i \in \{1,...,n_c\}}$, audio features $a_{i \in \{1,...,n_c\}}$, and query features $t_{i \in \{1,...,n_t\}}$ extracted by the corresponding pretrained model, CFSum first uses modal autoencoders to enhance the features of the corresponding modality under global information. The features are then input into the modal fusion module that performs coarse-grained fusion, yielding the fused representations $v^\text{fused}_{i \in \{1,...,n_c\}}$, $a^\text{fused}_{i \in \{1,...,n_c\}}$, and $t^\text{fused}_{i \in \{1,...,n_t\}}$. These representations are next input to the feature interaction module that performs fine-grained information interaction and query-relevance detection, generating $z^\text{out}_{i \in \{1,...,n_c\}}$. Finally, the latter is used by the saliency prediction head to calculate losses and obtain the clip-level saliency scores $\hat{s}_{i \in \{1,...,n_c\}}$.

\begin{table*}
\begin{floatrow}
\renewcommand\tabcolsep{0pt}
\footnotesize
\hspace{-0.21cm}
\vspace{-0.5em}
\ttabbox[0.565\textwidth]{
\caption{Results on TVSum. V and A denote video and audio modalities input (In) to the model. \textbf{Bold} and \underline{underline} denote first and second results.}
\vspace{-1em}}{
\label{tab1}
\begin{tabularx}{\linewidth}{@{\hspace{1mm}}p{1.9cm}>{\centering}p{0.5cm}>{\centering}p{0.7cm}<{\centering}p{0.7cm}<{\centering}p{0.7cm}<{\centering}p{0.7cm}<{\centering}p{0.7cm}<{\centering}p{0.7cm}<{\centering}p{0.7cm}<{\centering}p{0.7cm}<{\centering}p{0.7cm}<{\centering}p{0.7cm}<{\centering}p{0.7cm}<{\centering}}
\toprule
\textbf{Model} & \textbf{In} & \textbf{VT} & \textbf{VU} & \textbf{GA} & \textbf{MS} & \textbf{PK} & \textbf{PR} & \textbf{FM} & \textbf{BK} & \textbf{BT} & \textbf{DS} & \textbf{Avg.} \\
\midrule
sLSTM \cite{zhang2016video} & V & 41.1 & 46.2 & 46.3 & 47.7 & 44.8 & 46.1 & 45.2 & 40.6 & 47.1 & 45.5 & 45.1\\
LIM-S \cite{xiong2019less} & V & 55.9 & 42.9 & 61.2 & 54.0 & 60.3 & 47.5 & 43.2 & 66.3 & 69.1 & 62.6 & 56.3\\
SL-Module \cite{xu2021cross} & V & 86.5 & 68.7 & 74.9 & {86.2} & 79.0 & 63.2 & 58.9 & 72.6 & 78.9 & 64.0 & 73.3\\
UniVTG \cite{lin2023univtg} & V         & 83.9 & 85.1 & 89.0 & 80.1 & 84.6 & 81.4 & 70.9 & 91.7 & 73.5 & 69.3 & 81.0 \\
QD-DETR \cite{moon2023query}    & V     & 88.2 & 87.4 & 85.6 & {85.0} & 85.8 & 86.9 & 76.4 & 91.3 & 89.2 & 73.7 & 85.0 \\
R$^2$-Tuning \cite{liu2024r}    & V  & 85.0 & 85.9 & 91.0 & 81.7 & {88.8} & {87.4} & {78.1} & 89.2 & {90.3} & 74.7 & 85.2 \\
CG-DETR \cite{moon2023correlation}   & V      & 86.9 & {88.8} & {94.8} & {87.7} & {86.7} & {89.6} & 74.8 & {93.3} & 89.2 & 75.9 & \underline{86.8} \\
\textbf{CFSum}(Ours) & V & {86.8} & {84.4} & {92.6} & 79.6 & {83.9} & {79.2} & {75.5} & {94.1} & {86.9} & {79.4} & {84.3}\\
\midrule
MINI-Net \cite{hong2020mini} & VA & 80.6 & 68.3 & 78.2 & 81.8 & 78.1 & 65.8 & 57.8 & 75.0 & 80.2 & 65.5 & 73.2\\
TCG \cite{ye2021temporal} & VA & 85.0 & 71.4 & 81.9 & 78.6 & 80.2 & 75.5 & 71.6 & 77.3 & 78.6 & 68.1 & 76.8\\
Joint-VA \cite{badamdorj2021joint} & VA & 83.7 & 57.3 & 78.5 & {86.1} & 80.1 & 69.2 & 70.0 & 73.0 & {97.4} & 67.5 & 76.3\\
CO-AV \cite{li2022probing} & VA      & {90.8} & 72.8 & 84.6 & {85.0} & 78.3 & 78.0 & 72.8 & 77.1 & {89.5} & 72.3 & 80.1 \\
UMT \cite{liu2022umt} & VA & 87.5 & 81.5 & 88.2 & 78.8 & 81.4 & 87.0 & 76.0 & 86.9 & 84.4 & 79.6 & 83.1\\
UVCOM \cite{xiao2024bridging}  & VA    & 87.6 & {91.6} & 91.4 & {86.7} & {86.9} & 86.9 & {76.9} & {92.3} & 87.4 & 75.6 & {86.3} \\
\textbf{CFSum}(Ours) & VA & {89.9} & {88.2} & {93.8} & 84.7 & {86.1} & {91.2} & {78.1} & {87.2} & {86.6} & {83.0} & \textbf{86.9}\\
\bottomrule
\end{tabularx}}
\hspace{-0.15cm}
\vspace{-0.5em}
\ttabbox[0.412\textwidth]{
\caption{Results on Youtube Highlights. V and A denote video and audio modalities input (In) to the model.}
\vspace{-1em}}{
\label{tab2}
\begin{tabularx}{\linewidth}{@{\hspace{1mm}}p{1.9cm}>{\centering}p{0.5cm}>{\centering}p{0.7cm}<{\centering}p{0.7cm}<{\centering}p{0.7cm}<{\centering}p{0.7cm}<{\centering}p{0.7cm}<{\centering}p{0.7cm}<{\centering}p{0.7cm}<{\centering}}
\toprule
\textbf{Model} & \textbf{In} & \textbf{Dog} & \textbf{Gym} & \textbf{Par} & \textbf{Ska} & \textbf{Ski} & \textbf{Sur} & \textbf{Avg.} \\
\midrule
LSVM \cite{sun2014ranking} & V & 60.0 & 41.0 & 61.0 & 62.0 & 36.0 & 61.0 & 53.6 \\
LIM-S \cite{xiong2019less} & V & 57.9 & 41.7 & 67.0 & 57.8 & 48.6 & 65.1 & 56.4 \\
SL-Module \cite{xu2021cross} & V & {70.8} & 53.2 & 77.2 & {72.5} & 66.1 &76.2 & 69.3 \\
UniVTG \cite{lin2023univtg}   & V      & 71.8 & 76.5 & 73.9 & 73.3 & 73.2 & 82.2 & 75.2 \\
QD-DETR \cite{moon2023query}    & V     & 72.2 & {77.4} & 71.0 & 72.7 & 72.8 & 80.6 & 74.4 \\
R$^2$-Tuning \cite{liu2024r}  & V    & {75.6} & 73.5 & 73.0 & 74.6 & {74.8} & {84.8} & 76.1 \\
CG-DETR \cite{moon2023correlation}   & V      & {76.3} & 76.1 & 76.0 & 75.1 & {81.9} & 75.9 & 75.9 \\
\textbf{CFSum}(Ours) & V & 68.2 & {76.0} & {85.2} & {82.6} & {73.9} & {85.1} & \underline{78.5} \\
\midrule
MINI-Net \cite{hong2020mini} & VA & 58.2 & 61.7 & 70.2 & 72.2 & 58.7 & 65.1 & 64.4 \\
TCG \cite{ye2021temporal} & VA & 55.4 & 62.7 & 70.9 & 69.1 & 60.1 & 59.8 & 63.0 \\
Joint-VA \cite{badamdorj2021joint} & VA & 64.5 & 71.9 & 80.8 & 62.0 & {73.2} & 78.3 & 71.8 \\
CO-AV \cite{li2022probing}     & VA  & 60.9 & 66.0 & {89.0} & 74.1 & 69.0 & 81.1 & 74.7 \\
UMT \cite{liu2022umt} & VA & 65.9 & {75.2} & {81.6} & 71.8 & 72.3 & {82.7} & {74.9} \\
UVCOM \cite{xiao2024bridging}  & VA   & 73.8 & {77.1} & 75.7 & 75.3 & 74.0 & 82.7 & 76.4 \\
\textbf{CFSum}(Ours) & VA & {69.6} & {76.4} & {82.5} & {86.1} & {73.6} & {83.7} & \textbf{78.7} \\
\bottomrule
\end{tabularx}}
\end{floatrow}
\vspace{-1em}
\end{table*}

\begin{table}[ht]
    \centering
    \footnotesize
    \renewcommand{\arraystretch}{1}
    \setlength{\tabcolsep}{6pt}
    \caption{Results on QVHighlights. V and A denote video and audio modalities. \# of Params denotes number of parameters.}
    \vspace{-1em}
    \label{tab3}
    \scalebox{1}{
    \begin{tabular}{c c cc c}
    \toprule
        \multicolumn{1}{c}{\textbf{Model}} & \multicolumn{1}{c}{\textbf{Input}} & \multicolumn{2}{c}{\textbf{Saliency $\geq$ Very Good}} & \multicolumn{1}{c}{\textbf{\# of Params}} \\
        {} & {\textbf{Modality}} & {\ \ \ \ mAP} & {\ \ \ HIT@1} & {} \\
        \cmidrule(lr){1-1} \cmidrule(lr){2-2} \cmidrule(lr){3-4} \cmidrule(lr){5-5}
        QD-DETR \cite{moon2023query} & V & \ \ \ \ 38.94 & \ \ \ 62.40 & 7.6M \\
        MH-DETR \cite{xu2024mh} & V & \ \ \ \ 38.22 & \ \ \ 60.51 & 8.2M \\
        UniVTG \cite{lin2023univtg} & V & \ \ \ \ 38.20 & \ \ \ 60.96 & 41.3M \\
        TR-DETR \cite{sun2024tr} & V & \ \ \ \ 39.91 & \ \ \ 63.42 & 7.9M \\
        EaTR \cite{jang2023knowing} & V & \ \ \ \ 37.15 & \ \ \ 58.65 & 9.0M \\
        \midrule
        UMT \cite{liu2022umt} & VA & \ \ \ \ 38.18 & \ \ \ 59.99 & 14.9M \\
        UVCOM \cite{xiao2024bridging} & VA & \ \ \ \ 39.79 & \ \ \ 64.79 & - \\
        \textbf{CFSum}(Ours) & VA & \ \ \ \ \textbf{41.18} & \ \ \ \textbf{66.37}  & 12.9M \\
    \bottomrule
    \end{tabular}
    }
    \vspace{-2.5em}
\end{table}

\subsection{Modal Autoencoder}

Prior research \cite{liu2022umt, park2022multimodal} has revealed that the feature representations extracted from pretrained models fail to provide enough global contextual information crucial for video summarization. To augment our representations with global information, we adopt transformer-based \cite{vaswani2017attention} modal autoencoders that perform intra-modal learning. CFSum's modal autoencoder comprises multi-head self-attention layers, allowing for interaction among features within a modality. The input feature representation $f_i \in \{v_i, a_i, t_i\}$ simultaneously serves as the \textit{query}, \textit{key}, and \textit{value} for the self-attention computation. Mathematically,
\begin{align}
f^\text{att}_i &= \omega_v \softmax\!\left(\frac{\omega_q f_i \times (\omega_k f_j)^T}{\sqrt{d_k}}\right) f_i,\\
f^\text{aug}_i &= f_i + f^\text{att}_i,
\end{align}
where $f_{i, j \in \{1,...,n_c\}}$ if $f\equiv v$ or $f\equiv a$, or $f_{i, j \in \{1,...,n_t\}}$ if $f\equiv t$, $d_k$ is the feature dimensionality of $f_i$, and the \textit{query}, \textit{key}, \textit{value} weights are $\omega_q$, $\omega_k$, $\omega_v$. The augmented expression $f^\text{aug}_i$ is then projected into a unified space using an MLP.

\subsection{Modal Fusion Module}

As stated earlier, existing work focuses mainly on the fusion of video and text modalities, employing cross-attention operations on video-query pairs for information interaction, but it does not effectively leverage the unique information of each modality. To incorporate information from different modalities, CFSum starts by performing coarse-grained fusion in a modal fusion module to jointly model the visual-text-audio representation. The augmented feature representations $v^\text{aug}_i$, $a^\text{aug}_i$, and $t^\text{aug}_i$ are concatenated into a joint representation $z_{i \in \{1,...,n_c+n_c+n_t\}}$, which is then input to the modal fusion module that performs inter-modality modeling. Given the advantages of transformers in multi-modal learning, the modal fusion module has a multi-head self-attention structure, where the computation uses $z_i$ simultaneously as the \textit{query}, \textit{key}, and \textit{value}, as follows:
\begin{align}
z_i &= \concat(v^\text{aug}_i, a^\text{aug}_i, t^\text{aug}_i),\\
z^\text{att}_i &= \omega_v \softmax\!\left(\frac{\omega_q z_i \times (\omega_k z_j)^T}{\sqrt{d_k}}\right) z_i,\\ 
z^\text{fused}_i &= z_i + z^\text{att}_i,
\end{align}
where $i,j \in \{1,...,n_c+n_c+n_t\}$ and $z^\text{fused}_{i \in \{1,...,n_c+n_c+n_t\}}$ is the fused visual-text-audio representation.

Thus, the  modal fusion module enables global information interaction between the feature information of one modality and the other modalities, aggregating the multi-modal global contextual information into each modality segment.

Having computed $z^\text{fused}_i$, it is again split according to the original feature lengths of each modality, resulting in the fused video, audio, and text feature representations, denoted $v^\text{fused}_i$, $a^\text{fused}_{i \in \{1,...,n_c\}}$, and $t^\text{fused}_{i \in \{1,...,n_t\}}$, respectively. These representations are then input to a feed-forward network for further feature enhancement. This coarse-grained fusion stage enforces the feature interactions between modalities and outputs the attended feature representations of each modality.

\begin{table*}[h]
\centering
\caption{Contribution of each modality used in CFSum for TVSum and Youdtube Highlights. V, A, and T denote video, audio, and text.}
\vspace{-0.5em}
\resizebox{0.99\textwidth}{!}{
  \begin{tabular}{ccc ccccccccccc ccccccc}
\toprule
\multicolumn{3}{c}{\textbf{Modality}} & \multicolumn{11}{c}{\textbf{TVSum}} & \multicolumn{7}{c}{\textbf{Youtube Highlights}} \\
V & T & {A} & {VT} & {VU} & {GA} & {MS} & {PK} & {PR} & {FM} & {BK} & {BT} & {DS} & {Avg.} & {Dog} & {Gym} & {Par} & {Ska} & {Ski} & {Sur} & {Avg.} \\
\cmidrule(lr){1-3} \cmidrule(lr){4-14} \cmidrule(lr){15-21}
\checkmark & & & {86.8} & {84.4} & {92.6} & 79.6 & {83.9} & {79.2} & {75.5} & \textbf{94.1} & \textbf{86.9} & {79.4} & {84.3} & 68.2 & {76.0} & \textbf{85.2} & {82.6} & \textbf{73.9} & \textbf{85.1} & {78.5}\\ 
\checkmark & \checkmark & & {88.4} & {86.3} & {92.8} & 80.8 & {85.1} & {85.2} & {75.7} & {85.7} & {86.1} & {79.0} & {84.5} & -- & -- & -- & -- & -- & -- & -- \\
\checkmark & & \checkmark &  -- & -- & -- & -- & -- & -- & -- & -- & -- & -- & -- & \textbf{69.6} & \textbf{76.4} & {82.5} & \textbf{86.1} & {73.6} & {83.7} & \textbf{78.7} \\
\checkmark & \checkmark & \checkmark & \textbf{89.9} & \textbf{88.2} & \textbf{93.8} & \textbf{84.7} & \textbf{86.1} & \textbf{91.2} & \textbf{78.1} & 87.2 & {86.6} & \textbf{83.0} & \textbf{86.9} & -- & -- & -- & -- & -- & -- & --\\
\bottomrule
\end{tabular}}
\vspace{-2em}
\label{tab4}
\end{table*}

\begin{table}[t]
    \centering
    \footnotesize
    \renewcommand{\arraystretch}{1}
    \setlength{\tabcolsep}{6pt}
    \caption{Contribution of each module used in CFSum for QVHighlights. AE, MFM, and FIM denote Modal Autoencoder, Modal Fusion Module, and Feature Interaction Module.}
    \vspace{-1em}
    \label{tab5}
    \scalebox{1}{
    \begin{tabular}{ccc cc}
    \toprule
        \multicolumn{3}{c}{\textbf{Module}} & \multicolumn{2}{c}{\textbf{Saliency $\geq$ Very Good}} \\
        AE & MFM & FIM & {\ \ \ \ mAP} & {\ \ \ HIT@1} \\
        \cmidrule(lr){1-3} \cmidrule(lr){4-5}
        \checkmark &  &  & \ \ \ \ {24.53} & \ \ \ {33.23} \\
        \checkmark & \checkmark &  & \ \ \ \ {36.64} & \ \ \ {59.74} \\
        \checkmark &  & \checkmark & \ \ \ \ {36.56} & \ \ \ {59.48} \\
         & \checkmark & \checkmark & \ \ \ \ {39.96} & \ \ \ {64.55} \\
        \midrule
        \checkmark & \checkmark & \checkmark & \ \ \ \ \textbf{41.18} & \ \ \ \textbf{66.37}  \\
    \bottomrule
    \end{tabular}
    }
    \vspace{-2em}
\end{table}

\subsection{Feature Interaction Module}

Although the modal fusion module in the coarse-grained fusion stage ensures that each modality's features can interact with the other modalities, it inevitably introduces the background noise of the video and audio to the fused features. This noise hinders the identification of important or query-relevant clips in the video. Therefore, CFSum proceeds to perform fine-grained fusion by feature interaction modules that facilitate more detailed modality fusion mitigating the impact of noise. More specifically, the interaction module employs two multi-head cross-attention layers to capture text-video and text-audio interactions. The results are then weighted and summed to form the final inter-modality representation, as follows:
\begin{align}
tx^\text{att}_i &= \omega_v \softmax\!\left(\frac{\omega_q t^\text{fused}_i \times (\omega_k x^\text{fused}_i)^T}{\sqrt{d_k}}\right) x^\text{fused}_i, \\
z^\text{out}_i &= \sum \omega_\text{tx} tx^\text{att}_i,
\end{align}
where $x^\text{fused}_i \in \{v^\text{fused}_i, a^\text{fused}_i\}$ are the fused representations, $tx^\text{att}_{i \in \{1,...,n_c\}} \in \{tv^\text{att}_i$, $ta^\text{att}_i\}$ are the text-video and text-audio cross-attention outputs, $\omega_\text{tx} \in \{\omega_\text{tv}$, $\omega_\text{ta}\}$ are weights in the summation, and $z^\text{out}_i$ is the final inter-modality representation. This is fed to the prediction head to predict saliency scores.

\subsection{Saliency Prediction Head}

Representation $z^\text{out}_i$ is input into a linear layer that computes clip-level saliency scores $\hat{s_i}$, which are fed to a saliency prediction head that calculates the mean squared error losses
\begin{equation}
\mathcal{L} = \frac{1}{n_c} \sum^{n_c}_{i=1} (s_i - \hat{s_i})^2 \label{eq8}
\end{equation}
with respect to the ground truth saliency scores $s_i$ to optimize the learning process.

\section{Experiments}

\subsection{Dataset Description and Implementation Details}

\paragraph{Datasets} Following previous work \cite{liu2022umt, moon2023query, park2022multimodal}, we utilized the TVSum \cite{song2015tvsum}, YouTube Highlights \cite{sun2014ranking} and QVHighlights \cite{lei2107qvhighlights} datasets. TVSum contains 50 annotated videos divided into 10 categories with 5 videos in each category. Youtube Hightlights contains a total of 433 videos across 6 different categories. QVHighlights contains 10,148 videos, each has at least one text query. For all datasets, we followed the original data-split settings.

\paragraph{Implementation Details} For a fair comparison, we used the same pretrained models as in previous work \cite{liu2022umt}---i.e., I3D \cite{carreira2017quo}, PANN \cite{kong2020panns}, and CLIP \cite{radford2021learning} to extract video, audio, and text features, respectively. For model training, we used the Adam optimizer \cite{kingma2014adam} with $1 \times 10^{-3}$ learning rate and $1 \times 10^{-4}$ weight decay. We trained the models for 500 epochs for TVSum and YouTube Highlights, and 100 epochs for QVHighlights. In modal architecture, all CFSum attention layers had 8 heads and positional encodings. The weights $\omega_\text{tv}$, $\omega_\text{ta}$ were initially set to 2.0 and 1.0. For consistency with prior work, we used mean Average Precision (mAP) and HIT@1 as the evaluation metric.

Note that the audio data for all datasets can be obtained from the transcription of videos. We utilized video titles as the text input to TVSum. Not all videos in YouTube Highlights have titles, however, so for this dataset we input only video and audio to the model.

\subsection{Performance Comparison}

We compared the performance of CFSum against other methods on three datasets, as reported in \autoref{tab1}, \autoref{tab2} and \autoref{tab3}. The results consistently demonstrate that CFSum outperforms other methods across these datasets. With video, audio, and text inputs, our CFSum significantly reduces training parameters while achieving substantial performance improvements compared to previous methods, reaching SOTA performance. This demonstrates the effectiveness and conciseness of our approach. Compared to methods requiring only video and text inputs, our method introduces a small number of additional parameters but achieves a notable performance improvement, and supports audio modality input.

\subsection{Ablation Study}

To analyze the contribution of each modality and module, we present ablation studies results in \autoref{tab4} and \autoref{tab5}. \paragraph{Modality} CFSum shows a significant performance improvement with multi-modal input compared to single-modal input, further indicating that CFSum exploits the feature information inherent in each modality. Note that in certain video types, audio input can effectively enhance model performance, which highlights the importance of the audio modality.
\paragraph{Module} The results show that all modules, especially the Modal Fusion Module and Feature Interaction Module, can greatly promote information interaction between modalities, illustrating that the two-stage fusion structure of CFSum achieve deep fusion both within and between modalities.

\section{Conclusions}

We propose CFSum, a novel transformer-based multi-modal video summarization framework that employs coarse-fine fusion. CFSum adopts a two-stage feature fusion structure to model intra-modal and inter-modal fusion, fully utilizing the information inherent to each modality. Unlike prior work, CFSum emphasizes the unique information contained in the audio modality, elevating its importance to be commensurate with that of video and text. Our extensive experiments comparing the performances of CFSum and previous MMVS models have confirmed the effectiveness and superiority of CFSum.


\balance
\bibliographystyle{IEEEtran}
\bibliography{IEEEabrv,cfsum}

\end{document}